%% file: emnlp2018.tex
\documentclass[11pt,a4paper]{article}
\usepackage[hyperref]{emnlp2018}
\usepackage{times}
\usepackage{latexsym}
\usepackage{graphicx}
\usepackage{CJKutf8}
\usepackage{bm}
\usepackage{xcolor}
\usepackage{amsmath,amsthm,amssymb,amsfonts}

\usepackage{url}

\aclfinalcopy 

\title{Language Modeling with Sparse Product of Sememe Experts}

\author{Yihong Gu$^{1,2,}\thanks{Equal contribution.}$ $ $ $ $\quad Jun Yan$^{1,3,*}$\quad Hao Zhu$^{1,2,*}$\quad Zhiyuan Liu$^{1,2,}$\thanks{Correspondence author.}\\
\textbf{Ruobing Xie$^4$\quad Maosong Sun$^{1,2}$\quad Fen Lin$^4$\quad Leyu Lin$^4$}  \\
  $^1$Institute for Artificial Intelligence\\
  State Key Lab on Intelligent Technology and Systems\\
  $^2$Department of CST, $^3$Department of EE, Tsinghua University, Beijing, China\\
  $^4$Search Product Center, WeChat Search Application Department, Tencent\\
  {\tt \{gyh15,j-yan15,zhuhao15\}@mails.tsinghua.edu.cn}, \\
  {\tt \{lzy,sms\}@tsinghua.edu.cn}, 
  {\tt xrbsnowing@163.com}, \\
  {\tt \{felicialin,goshawklin\}@tencent.com}\\
  }

\usepackage{scrextend}
\deffootnote[1.5em]{1.5em}{1em}{\thefootnotemark\space}

\date{}

\begin{document}
\maketitle

\input{abstract}
\input{introduction}
\input{background}
\input{methodology}
\input{experiments}
\input{relatedwork}
\input{conclusion}
\input{acknowledgement}
\bibliography{emnlp2018}
\bibliographystyle{acl_natbib_nourl}
\clearpage
\newpage
\newpage

\appendix
\input{appendix}

\end{document}

%% file: abstract.tex
\begin{abstract}
Most language modeling methods rely on large-scale data to statistically learn the sequential patterns of words. In this paper, we argue that words are atomic language units but not necessarily atomic semantic units. Inspired by HowNet, we use sememes, the minimum semantic units in human languages, to represent the implicit semantics behind words for language modeling, named Sememe-Driven Language Model (SDLM). More specifically, to predict the next word, SDLM first estimates the sememe distribution given textual context. Afterwards, it regards each sememe as a distinct semantic expert, and these experts jointly identify the most probable senses and the corresponding word. In this way, SDLM enables language models to work beyond word-level manipulation to fine-grained sememe-level semantics, and offers us more powerful tools to fine-tune language models and improve the interpretability as well as the robustness of language models. Experiments on language modeling and the downstream application of headline generation demonstrate the significant effectiveness of SDLM. Source code and data used in the experiments can be accessed at \url{https://github.com/thunlp/SDLM-pytorch}.
\end{abstract}

%% file: introduction.tex
\section{Introduction}
Language Modeling (LM) aims to measure the probability of a word sequence, reflecting its fluency and likelihood as a feasible sentence in a human language. Language Modeling is an essential component in a wide range of natural language processing (NLP) tasks, such as Machine Translation \cite{brown1990statistical,brants2007large}, Speech Recognition \cite{katz1987estimation}, Information Retrieval \cite{berger1999information,ponte1998language,miller1999hidden,hiemstra1998linguistically} and Document Summarization \cite{rush2015neural,banko2000headline}.

A probabilistic language model calculates the conditional probability of the next word given their contextual words, which are typically learned from large-scale text corpora. Taking the simplest language model for example, N-Gram estimates the conditional probabilities according to maximum likelihood over text corpora \cite{jurafsky2000speech}. Recent years have also witnessed the advances of Recurrent Neural Networks (RNNs) as the state-of-the-art approach for language modeling \cite{mikolov2010recurrent}, in which the context is represented as a low-dimensional hidden state to predict the next word.

\begin{figure}[!t]
\includegraphics[width = .475\textwidth]{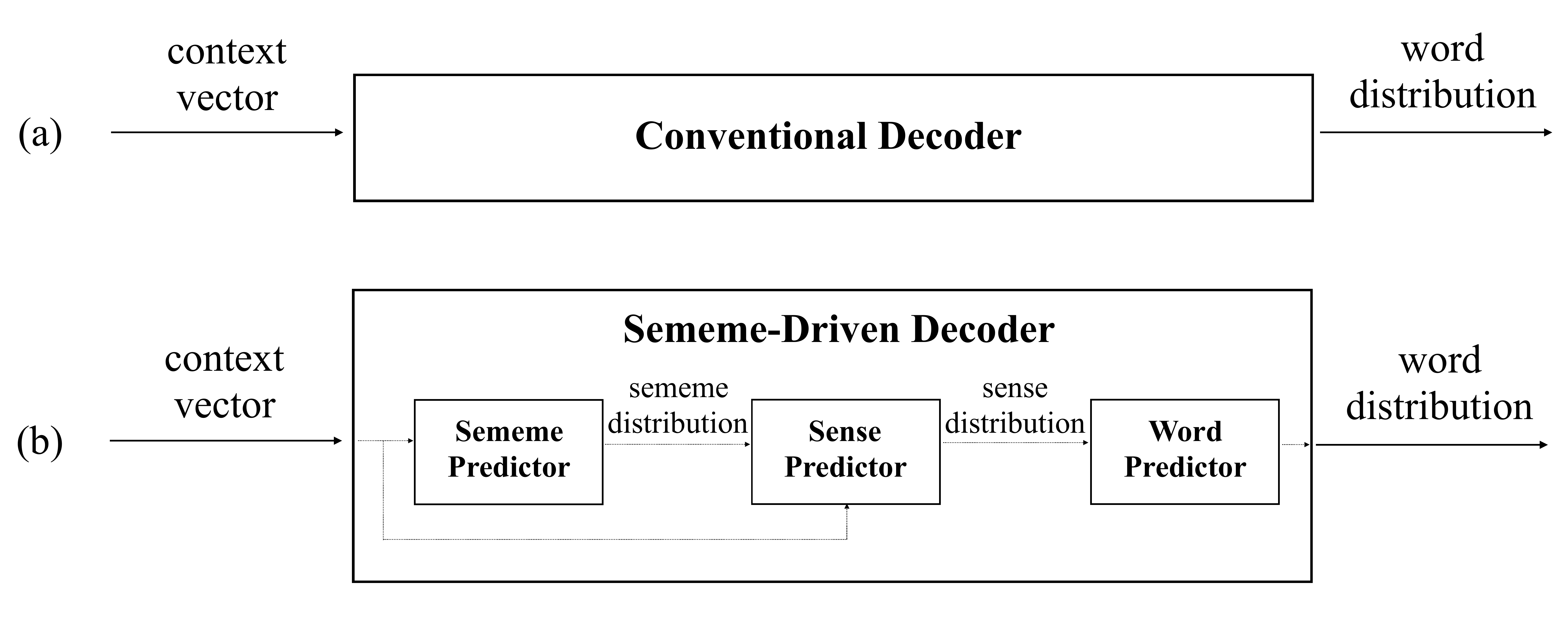}
\caption{Decoder of (a) Conventional Language Model, (b) Sememe-Driven Language Model.}
\label{fig:task_demo}
\end{figure}

Those conventional language models including neural models typically assume words as atomic symbols and model sequential patterns at word level. However, this assumption does not necessarily hold to some extent. Let us consider the following example sentence for which people want to predict the next word in the blank, 
\begin{center}
\textit{The U.S. trade deficit last year is initially estimated to be 40 billion \underline {\hbox to 10mm{}} .}
\end{center}
People may first realize a \emph{unit} should be filled in, then realize it should be a \emph{currency unit}. Based on the country this sentence is talking about, the U.S., one may confirm it should be an \emph{American currency unit} and predict the word \emph{dollars}. Here, the \emph{unit}, \emph{currency}, and \emph{American} can be regarded as basic semantic units of the word \emph{dollars}. This process, however, has not been explicitly taken into consideration by conventional language models. That is, although in most cases words are atomic language units, words are not necessarily atomic semantic units for language modeling. We argue that explicitly modeling these atomic semantic units could improve both the performance and the interpretability of language models.

Linguists assume that there is a limited close set of atomic semantic units composing the semantic meanings of an open set of concepts (i.e. word senses). These atomic semantic units are named \textbf{sememes}~\citep{zhendong2006hownet}.\footnote{Note that although sememes are defined as the minimum semantic units, there still exist several sememes for capturing syntactic information. For example, the word \begin{CJK}{UTF8}{gbsn}和\end{CJK} ``with'' corresponds to one specific sememe \begin{CJK}{UTF8}{gbsn}功能词\end{CJK} ``FunctWord''.} Since sememes are naturally implicit in human languages, linguists have devoted much effort to explicitly annotate lexical sememes for words and build linguistic common-sense knowledge bases. HowNet~\citep{zhendong2006hownet} is one of the representative sememe knowledge bases, which annotates each Chinese word sense with its sememes. The philosophy of HowNet regards the parts and attributes of a concept can be well represented by sememes. HowNet has been widely utilized in many NLP tasks such as word similarity computation \citep{liu2002word} and sentiment analysis \citep{xianghua2013multi}. However, less effort has been devoted to exploring its effectiveness in language models, especially neural language models.

It is non-trivial for neural language models to incorporate discrete sememe knowledge, as it is not compatible with continuous representations in neural models. In this paper, we propose a Sememe-Driven Language Model (SDLM) to leverage lexical sememe knowledge. In order to predict the next word, we design a novel sememe-sense-word generation process: (1) We first estimate sememes' distribution according to the context. (2) Regarding these sememes as experts, we propose a sparse product of experts method to select the most probable senses. (3) Finally, the distribution of words could be easily calculated by marginalizing out the distribution of senses.

We evaluate the performance of SDLM on the language modeling task using a Chinese newspaper corpus People's Daily \footnote{http://paper.people.com.cn/rmrb/} (Renmin Ribao), and also on the headline generation task using the Large Scale Chinese Short Text Summarization (LCSTS) dataset \cite{hu2015lcsts}. Experimental results show that SDLM outperforms all those data-driven baseline models. We also conduct case studies to show that our model can effectively predict relevant sememes given context, which can improve the interpretability and robustness of language models.

%% file: background.tex
\section{Background}

Language models target at learning the joint probability of a sequence of words $P(w^1, w^2, \cdots, w^n)$, which is usually factorized as $P(w^1, w^2, \cdots, w^n)=\prod_{t=1}^{n}{P(w^t|w^{<t})}$. \citet{neural-lm} propose the first Neural Language Model as a feed-forward neural network. \citet{mikolov2010recurrent} use RNN and a softmax layer to model the conditional probability. To be specific, it can be divided into two parts in series. First, a context vector $\mathbf{g}^t$ is derived from a deep recurrent neural network. Then, the probability $P(w^{t+1}|w^{\le t})=P(w^{t+1};\mathbf{g}^{t})$ is derived from a linear layer followed by a softmax layer based on $\mathbf{g}^{t}$. Let $\mathrm{RNN}(\cdot,\cdot;\bm{\theta}_{\mathrm{NN}})$ denote the deep recurrent neural network, where $\bm{\theta}_\mathrm{NN}$ denotes the parameters. The first part can be formulated as
\begin{equation}
\mathbf{g}^t=\mathrm{RNN} (\mathbf{x}_{w^t}, \{\mathbf{h}^{t-1}_{l}\}_{l=1}^L;\bm{\theta}_\mathrm{NN}).
\end{equation}
Here we use subscripts to denote layers and superscripts to denote timesteps. Thus $\mathbf{h}^t_l$ represents the hidden state of the $L$-th layer at timestep $t$. $\mathbf{x}_{w^t}\in\mathbb{R}^{H_0}$ is the input embedding of word $w^t$ where $H_0$ is the input embedding size. We also have $\mathbf{g}^t \in \mathbb{R}^{H_1}$, where $H_1$ is the dimension of the context vector. 

Supposing that there are $N$ words in the language we want to model, the second part can be written as
\begin{equation}
P(w^{t+1};\mathbf{g}^{t})=\frac{\exp({\mathbf{g}^t}^\mathrm{T} \mathbf{w}_{w^{t+1}})}{\sum_{w'}{\exp({\mathbf{g}^t}^\mathrm{T} \mathbf{w}_{w'})}},
\end{equation}
where $\mathbf{w}_w$ is the output embedding of word $w$ and $\mathbf{w}_1, \mathbf{w}_2, \cdots \mathbf{w}_N \in \mathbb{R}^{H_2}$. Here $H_2$ is the output embedding size. For a conventional neural language model, $H_2$ always equals to $H_1$. 

Given the corpus $\{w^t\}_{t=1}^n$, the loss function is defined by the negative log-likelihood: 
\begin{equation}
	\mathcal{L}(\bm{\theta})=-\frac{1}{n}\sum_{t=1}^n \log P(w^{t}|w^{<t};\bm{\theta}),
\end{equation}
where $\bm{\theta}=\{\{\mathbf{x}_i\}_{i=1}^N, \{\mathbf{w}_i\}_{i=1}^N, \bm{\theta}_{\mathrm{NN}}\}$ is the set of parameters that are needed to be trained.

%% file: methodology.tex
\section{Methodology}
In this section, we present our SDLM which utilizes sememe information to predict the probability of the next word. SDLM is composed of three modules in series: Sememe Predictor, Sense Predictor and Word Predictor. The Sememe Predictor first takes the context vector as input and assigns a weight to each sememe. Then each sememe is regarded as an expert and makes predictions about the probability distribution over a set of senses in the Sense Predictor. Finally, the probability of each word is obtained in the Word Predictor.

\begin{figure}
\centering
\includegraphics[width = .5\textwidth]{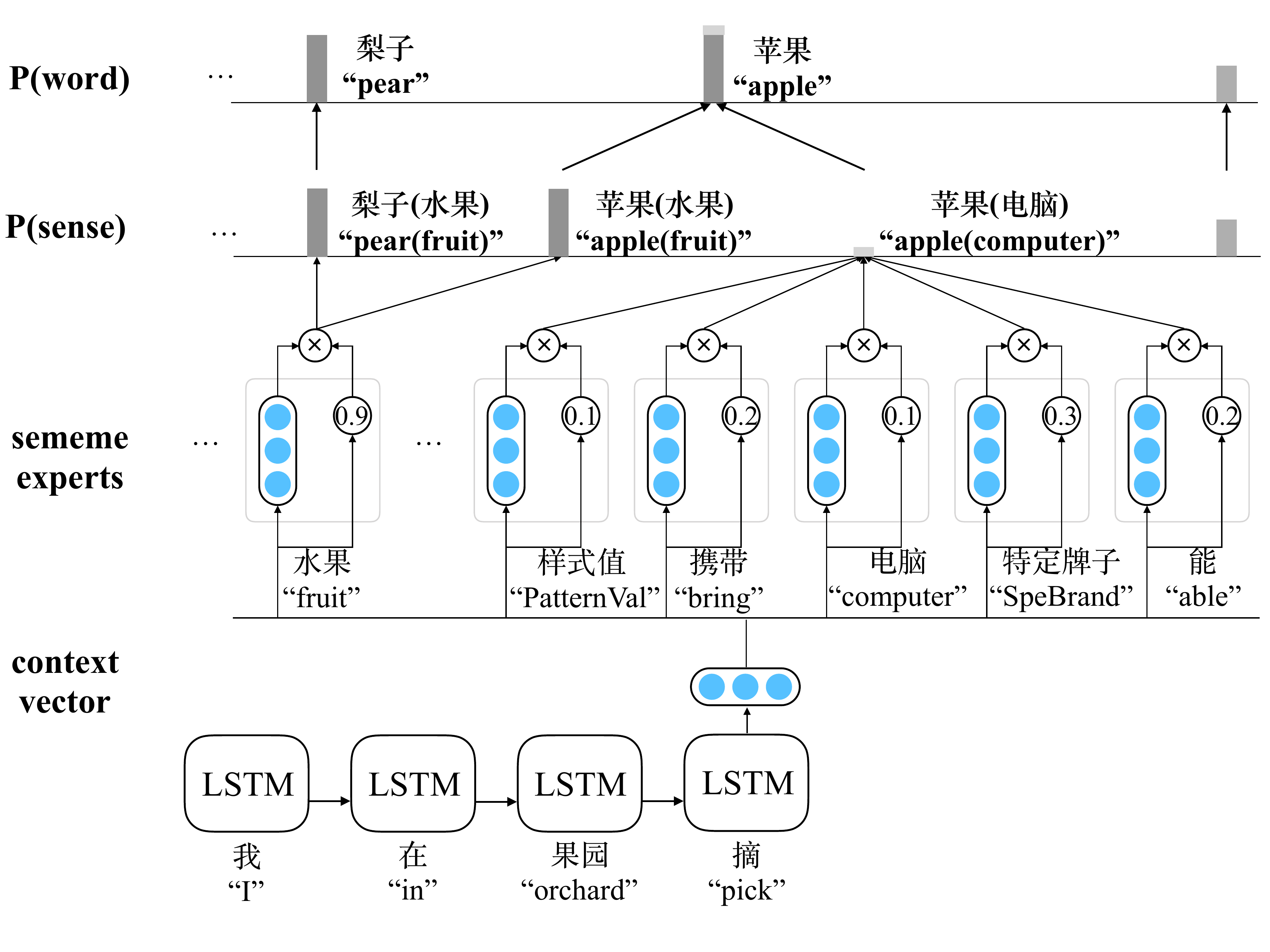}
\caption{An example of the architecture of our model.}
\label{architecture}
\end{figure}

Here we use an example shown in Figure \ref{architecture} to illustrate our architecture. Given context \begin{CJK}{UTF8}{gbsn}我在果园摘\end{CJK} ``In the orchard, I pick'', the actual next word could be \begin{CJK}{UTF8}{gbsn}苹果\end{CJK} ``apples''. From the context, especially the word \begin{CJK}{UTF8}{gbsn}果园\end{CJK} ``orchard'' and \begin{CJK}{UTF8}{gbsn}摘 ``pick''\end{CJK}, we can infer that the next word probably represents a kind of fruit. So the Sememe Predictor assigns a higher weight to the sememe \begin{CJK}{UTF8}{gbsn}水果\end{CJK} ``fruit'' (0.9) and lower weights to irrelevant sememes like \begin{CJK}{UTF8}{gbsn}电脑\end{CJK} ``computer'' (0.1). Therefore in the Sense Predictor, the sense \begin{CJK}{UTF8}{gbsn}苹果 (水果)\end{CJK} ``apple (fruit)'' is assigned a much higher probability than the sense \begin{CJK}{UTF8}{gbsn}苹果 (电脑)\end{CJK} ``apple (computer)''. Finally, the probability of the word \begin{CJK}{UTF8}{gbsn}苹果\end{CJK} ``apple'' is calculated as the sum of the probabilities of its senses \begin{CJK}{UTF8}{gbsn}苹果 (水果)\end{CJK} ``apple(fruit)'' and \begin{CJK}{UTF8}{gbsn}苹果 (电脑)\end{CJK} ``apple (computer)''.
 
In the following subsections, we first introduce the word-sense-sememe hierarchy in HowNet, and then give details about our SDLM.

\subsection{Word-Sense-Sememe Hierarchy}

\begin{figure}
\includegraphics[width = .5\textwidth]{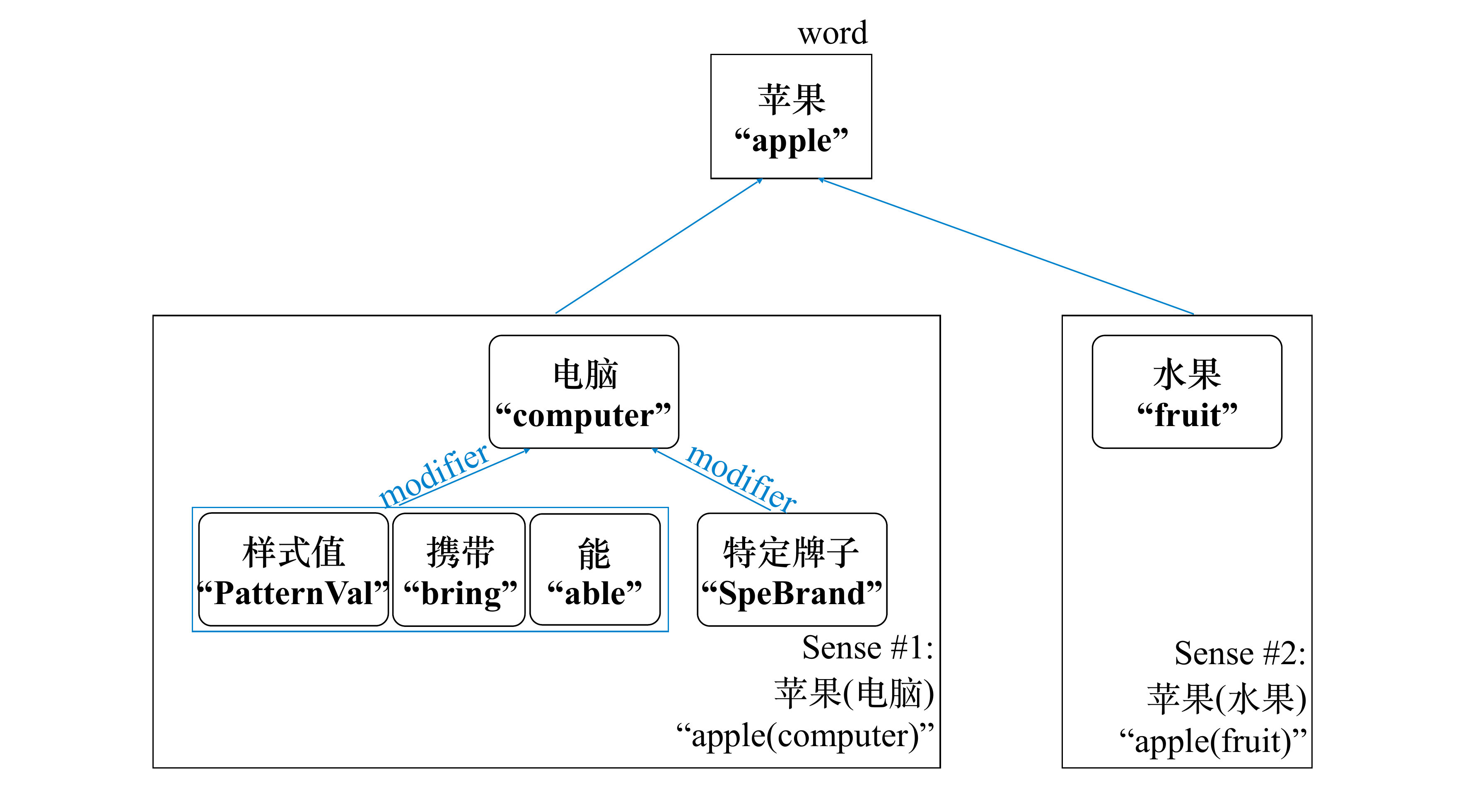}
\caption{An example of the word-sense-sememe hierarchy.}
\label{sememe_sample}
\end{figure}

We also use the example of ``apple'' to illustrate the word-sense-sememe hierarchy. As shown in Figure \ref{sememe_sample}, the word \begin{CJK}{UTF8}{gbsn}苹果\end{CJK} ``apple'' has two senses, one is the Apple brand, the other is a kind of fruit. Each sense is annotated with several sememes organized in a hierarchical structure. More specifically, in HowNet, sememes ``PatternVal'', ``bring'', ``SpeBrand'', ``computer'' and ``able'' are annotated with the word ``apple'' and organized in a tree structure. In this paper, we ignore the structural relationship between sememes. For each word, we group all its sememes as an unordered set.

We present the notations that we use in the following subsections as follows. We define the overall sememe, sense, and word set as $\mathcal{E}$, $\mathcal{S}$ and $\mathcal{W}$. And we suppose the corpus contains $K=|\mathcal{E}|$ sememes, $M=|\mathcal{S}|$ senses and $N=|\mathcal{W}|$ words. For word $w \in \mathcal{W}$, we denote its corresponding sense set as $\mathcal{S}^{(w)}$. For sense $s \in \mathcal{S}^{(w)}$, we denote its corresponding sememes as an unordered set $\mathcal{E}^{(s)}=\{e_{n_{1}},e_{n_{2}},\cdots,e_{n_{k}}\}\subset \mathcal{E}=\{e_k\}_{k=1}^K$.

\subsection{Sememe Predictor}

The Sememe Predictor takes the context vector $\mathbf{g} \in \mathbb{R}^{H_1}$ as input and assigns a weight to each sememe. We assume that given the context $w^{1}, w^{2}, \cdots, w^{t-1}$, the events that word $w^t$ contains sememe $e_k$ ($k \in \{1,2, \cdots, K\}$) are independent, since the sememe is the minimum semantic unit and there is no semantic overlap between any two different sememes. For simplicity, we ignore the superscript $t$. We design the Sememe Predictor as a linear decoder with the sigmoid activation function. Therefore, $q_k$, the probability that the next word contains sememe $e_k$, is formulated as
\begin{equation}
\small
	q_k = P(e_k|\mathbf{g}) = \sigma(\mathbf{g}^\mathrm{T} \mathbf{v}_k + b_k),
\end{equation}
where $\mathbf{v}_k \in \mathbb{R}^{H_1}$, $b_k \in \mathbb{R}$ are trainable parameters, and $\sigma(\cdot)$ denotes the sigmoid activation function. 

\subsection{Sense Predictor and Word Predictor}

The architecture of the Sense Predictor is motivated by Product of Experts (PoE) \citep{Hinton1999Products}. We regard each sememe as an expert that only makes predictions on the senses connected with it. Let $\mathcal{D}^{(e_k)}$ denote the set of senses that contain sememe $e_k$, the $k$-th expert. Different from conventional neural language models, which directly use the inner product of the context vector $\mathbf{g} \in \mathbb{R}^{H_1}$ and the output embedding $\mathbf{w}_w \in \mathbb{R}^{H_2}$ for word $w$ to generate the score for each word, we use $\phi^{(k)}(\mathbf{g}, \mathbf{w})$ to calculate the score given by expert $e_k$. And we choose a bilinear function parameterized with a matrix $\mathbf{U}_k \in \mathbb{R}^{H_1\times H_2}$ as a straight implementation of $\phi^{(k)}(\cdot, \cdot)$:
\begin{equation}
\small
\label{bilinear}
	\phi^{(k)}(\mathbf{g},\mathbf{w})=\mathbf{g}^\mathrm{T} \mathbf{U}_k \mathbf{w}.
\end{equation}

Let $\mathbf{w}_{s}$ denote the output embedding of sense $s$. The score of sense $s$ provided by sememe expert $e_k$ can be written as $\phi^{(k)}(\mathbf{g}, \mathbf{w}_{s})$. Therefore, $P^{(e_k)}(s|\mathbf{g})$, the probability of sense $s$ given by expert $e_k$, is formulated as
\begin{equation}
\small 
\label{single_expert}
	P^{(e_k)}(s|\mathbf{g}) = \frac{\exp(q_k C_{k,s}\phi^{(k)}(\mathbf{g}, \mathbf{w}_{s}))}{\sum_{s' \in \mathcal{D}^{(e_k)}}{\exp(q_k C_{k,s'}\phi^{(k)}(\mathbf{g}, \mathbf{w}_{s'}))}},
\end{equation}
where $C_{k,s}$ is a normalization constant because sense $s$ is not connected to all experts (the connections are sparse with approximately $\lambda N$ edges, $\lambda < 5$). Here we can choose either $C_{k,s} = 1/|\mathcal{E}^{(s)}|$ (\emph{left normalization}) or $C_{k,s} = 1/\sqrt{|\mathcal{E}^{(s)}||\mathcal{D}^{(e_k)}|}$ (\emph{symmetric normalization}).

In the Sense Predictor, $q_k$ can be viewed as a gate which controls the magnitude of the term $C_{k,s}\phi^{(k)}(\mathbf{g}, \mathbf{w}_{w_s})$, thus control the flatness of the sense distribution provided by sememe expert $e_k$. Consider the extreme case when $q_k \to 0$, the prediction will converge to the discrete uniform distribution. Intuitively, it means that the sememe expert will refuse to provide any useful information when it is not likely to be related to the next word.

Finally, we summarize the predictions on sense $s$ by taking the product of the probabilities given by relevant experts and then normalize the result; that is to say, $P(s|\mathbf{g})$, the probability of sense $s$, satisfies 
\begin{equation}
\small 
    P(s|\mathbf{g}) \propto \prod_{e_k \in \mathcal{E}^{(s)}}{P^{(e_k)}(s|\mathbf{g})}.
\end{equation}

Using Equation \ref{bilinear} and \ref{single_expert}, we can formulate $P(s|\mathbf{g})$ as
\begin{equation}
\small
 P(s|\mathbf{g}) = \frac{\exp(\sum_{e_k \in \mathcal{E}^{(s)}} q_k C_{k,s} \mathbf{g}^\mathrm{T} \mathbf{U}_k \mathbf{w}_{s})}{\sum_{s'}\exp(\sum_{e_k \in \mathcal{E}^{(s')}} q_k C_{k,s'} \mathbf{g}^\mathrm{T} \mathbf{U}_k \mathbf{w}_{s'})}.
\end{equation}

It should be emphasized that all the supervision information provided by HowNet is embodied in the connections between the sememe experts and the senses. If the model wants to assign a high probability to sense $s$, it must assign a high probability to some of its relevant sememes. If the model wants to assign a low probability to sense $s$, it can assign a low probability to its relevant sememes. Moreover, the prediction made by sememe expert $e_k$ has its own tendency because of its own $\phi^{(k)}(\cdot, \cdot)$. Besides, the sparsity of connections between experts and senses is also determined by HowNet itself. For our dataset, on average, a word is connected with 3.4 sememe experts and each sememe expert will make predictions about 22 senses.

As illustrated in Figure \ref{architecture}, in the Word Predictor, we get $P(w|\mathbf{g})$, the probability of word $w$, by summing up probabilities of corresponding $s$ given by the Sense Predictor, that is
\begin{equation}
\small 
    P(w|\mathbf{g}) = \sum_{s \in \mathcal{S}^{(w)}}{P(s|\mathbf{g})}.
\end{equation}

\subsection{Implementation Details}

\noindent \textbf{Basis Matrix} Actually, HowNet contains $K\approx 2000$ sememes. In practice, we cannot directly introduce $K \times H_1 \times H_2$ parameters, which might be computationally infeasible and lead to overfitting. To address this problem, we apply a weight-sharing trick called the basis matrix. We use $R$ basis matrices and their weighted sum to estimate $\mathbf{U}_k$: 

\begin{equation}
\small 
	\mathbf{U}_k = \sum_{r=1}^R{\alpha_{k,r} \mathbf{Q}_r} \label{basis_factor},
\end{equation}
where $\mathbf{Q}_r \in \mathbb{R}^{H_1\times H_2}$, $\alpha_{k,r} > 0$ are trainable parameters, and $\sum_{r=1}^R{\alpha_{k,r}}=1$.

\noindent \textbf{Weight Tying} To incorporate the weight tying strategy \citep{weight-tying,weight-tying-2}, we use the same output embedding for multiple senses of a word. To be specific, the sense output embedding $\mathbf{w}_s$ for each $s \in S^{(w)}$ is the same as the word input embedding $\mathbf{x}_w$.

%% file: experiments.tex
\section{Experiments}

We evaluate our SDLM on a Chinese language modeling dataset, namely People's Daily based on perplexity.\footnote{ Although we only conduct experiments on Chinese corpora, we argue that this model has the potential to be applied to other languages in the light of works on construction sememe knowledge bases for other languages, such as \citep{qi2018cross}.} Furthermore, to show that our SDLM structure can be a generic Chinese word-level decoder for sequence-to-sequence learning, we conduct a Chinese headline generation experiment on the LCSTS dataset. Finally, we explore the interpretability of our model with cases, showing the effectiveness of utilizing sememe knowledge.

\subsection{Language Modeling}

\subsubsection*{Dataset}

We choose the People's Daily Corpus, which is widely used for Chinese NLP tasks, as the resource. It contains one month’s news text from People’s Daily (Renmin Ribao). Taking Penn Treebank (PTB) \citep{penn} as a reference, we build a dataset for Chinese language modeling based on the People's Daily Corpus with 734k, 10k and 19k words in the training, validation and test set. After the preprocessing similar to \citep{mikolov2010recurrent} (see Appendix \ref{suc:preprocess_lm}), we get our dataset and the final vocabulary size is 13,476.

\subsubsection*{Baseline}

As for baselines, we consider three kinds of neural language modeling architectures with LSTM cells: simple LSTM, Tied LSTM and AWD-LSTM.

\noindent \textbf{LSTM and Tied LSTM} \citet{rnn-dropout} use the dropout strategy to prevent overfitting for neural language models and adopt it to two-layer LSTMs with different embedding and hidden size: 650 for medium LSTM, and 1500 for large LSTM. Employing the weight tying strategy, we get Tied LSTM with better performance. We set LSTM and Tied LSTM of medium and large size as our baseline models and use the code from PyTorch examples\footnote{\url{https://github.com/pytorch/examples/tree/master/word\_language\_model}} as their implementations.

\noindent \textbf{AWD-LSTM} 
Based on several strategies for regularizing and optimizing LSTM-based language models, \citet{awd-lstm-lm} propose AWD-LSTM as a three-layer neural network, which serves as a very strong baseline for word-level language modeling. We build it with the code released by the authors\footnote{\url{https://github.com/salesforce/awd-lstm-lm}}.

\noindent \textbf{Variants of Softmax} Meanwhile, to compare our SDLM with other language modeling decoders, we set cHSM (Class-based Hierarchical Softmax) \citep{Goodman2001Classes}, tHSM (Tree-based Hierarchical Softmax) \citep{Mikolov2013Distributed} and MoS (Mixture of Softmaxes) \citep{Yang2017Breaking} as the baseline add-on structures to the architectures above. 

\subsubsection*{Experimental Settings}

We apply our SDLM and other variants of softmax structures to the architectures mentioned above: LSTM (medium / large), Tied LSTM (medium / large) and AWD-LSTM. MoS and SDLM are only applied on the models that incorporate weight tying, while tHSM is only applied on the models without weight tying, since it is not compatible with this strategy.

For a fair comparison, we train these models with same experimental settings and conduct a hyper-parameter search for baselines as well as our models (the search setting and the optimal hyper-parameters can be found in Appendix \ref{sec:search_lm}). We keep using these hyper-parameters in our SDLM for all architectures. It should be emphasized that we use the SGD optimizer for all architectures, and we decrease the learning rate by a factor of 2 if no improvement is observed on the validation set. We uniformly initialize the word embeddings, the class embeddings for cHSM and the non-leaf embeddings for tHSM in $[-0.1, 0.1]$. In addition, we set $R$, the number of basis matrices, to $5$ in Tied LSTM architecture and to $10$ in AWD-LSTM architecture. We choose the \emph{left normalization} strategy because it performs better.

\subsubsection*{Experimental Results}

Table \ref{lm-result} shows the perplexity on the validation and test set of our models and the baseline models. From Table \ref{lm-result}, \ref{sense_tb}, and \ref{mean_sememe_tb}, we can observe that:

\noindent 1. Our models outperform the corresponding baseline models of all structures, which indicates the effectiveness of our SDLM. Moreover, our SDLM not only consistently outperforms state-of-the-art MoS model, but also offers much better interpretability (as described in Sect. \ref{sec:case study}), which makes it possible to interpret the prediction process of the language model. Note that under a fair comparison, we do not see MoS's improvement over AWD-LSTM while our SDLM outperforms it by 1.20 with respect to perplexity on the test set.

\noindent 2. To further locate the performance improvement of our SDLM, we study the perplexity of the single-sense words and multi-sense words separately on Tied LSTM (medium) and Tied LSTM (medium) + SDLM. Improvements with respect to perplexity are presented in Table \ref{sense_tb}. The performance on both single-sense words and multi-sense words gets improved while multi-sense words benefit more from SDLM structure because they have richer sememe information. 

\noindent 3. $ $ In Table \ref{mean_sememe_tb} we study the perplexity of words with different mean number of sememes. We can see that our model outperforms baselines in all cases and is expected to benefit more as the mean number of sememes increases.

\begin{table}
\centering \small
\begin{tabular}{lrrr}
  \hline
  \hline
  Model & \#Paras & Validation & Test\\
  \hline
  LSTM (medium) & 24M & 116.46 & 115.51 \\
  $\quad$ + cHSM & 24M & 129.12 & 128.12 \\
  $\quad$ + tHSM & 24M & 151.00 & 150.87 \\
  Tied LSTM (medium) & 15M & 105.35 & 104.67 \\
  $\quad$ + cHSM & 15M & 116.78 & 115.66 \\
  $\quad$ + MoS & 17M & 98.47 & 98.12 \\
  $\quad$ + SDLM & 17M & \textbf{97.75} & \textbf{97.32} \\
  \hline
  LSTM (large) & 76M & 112.39 & 111.66 \\
  $\quad$ + cHSM & 76M & 120.07 & 119.45 \\
  $\quad$ + tHSM & 76M & 140.41 & 139.61 \\
  Tied LSTM (large) & 56M & 101.46 & 100.71 \\
  $\quad$ + cHSM & 56M & 108.28 & 107.52 \\
  $\quad$ + MoS & 67M & 94.91 & 94.40 \\
  $\quad$ + SDLM & 67M & \textbf{94.24} & \textbf{93.60} \\
  \hline
  AWD-LSTM\footnotemark[4] & 26M & 89.35 & 88.86 \\
  $\quad$ + MoS & 26M & 92.98 & 92.76 \\
  $\quad$ + SDLM & 27M & \textbf{88.16} & \textbf{87.66} \\
  \hline
  \hline
\end{tabular}
\caption{Single model perplexity on validation and test sets on the People's Daily dataset.}
\label{lm-result}
\end{table}
\footnotetext[4]{We find that multi-layer AWD-LSTM has problems converging when adopting cHSM, so we skip that result.}
\setcounter{footnote}{4}

\noindent 1. Our models outperform the corresponding baseline models of all structures, which indicates the effectiveness of our SDLM. Moreover, our SDLM not only consistently outperforms state-of-the-art MoS model, but also offers much better interpretability (as described in Sect. \ref{sec:case study}), which makes it possible to interpret the prediction process of the language model. Note that under a fair comparison, we do not see MoS's improvement over AWD-LSTM while our SDLM outperforms it by 1.20 with respect to perplexity on the test set.

\noindent 2. To further locate the performance improvement of our SDLM, we study the perplexity of the single-sense words and multi-sense words separately on Tied LSTM (medium) and Tied LSTM (medium) + SDLM. Improvements with respect to perplexity are presented in Table \ref{sense_tb}. The performance on both single-sense words and multi-sense words gets improved while multi-sense words benefit more from SDLM structure because they have richer sememe information. 

\noindent 3. $ $ In Table \ref{mean_sememe_tb} we study the perplexity of words with different mean number of sememes. We can see that our model outperforms baselines in all cases and is expected to benefit more as the mean number of sememes increases.

\begin{table}[t]
\centering \small
\begin{tabular}{lrr}
   \hline
   \hline
   & \#senses $=$ 1 & \#senses $>$ 1 \\
   \hline
   Baseline ppl & 93.21 & 121.18 \\
   SDLM ppl & 87.22 & 111.88 \\
   $\Delta$ppl & 5.99 & 9.29 \\
   $\Delta$ppl/Baseline ppl & 6.4\% & 7.8\% \\
   \hline
   \hline
\end{tabular}
\caption{Perplexity of words with different number of senses on the test set.}
\label{sense_tb}
\end{table}

\begin{table}
\centering \small
\begin{tabular}{lrrrr}
   \hline
   \hline
   & [1, 2) & [2, 4) & [4, 7) & [7, 14) \\
   \hline
   Baseline ppl & 71.56 & 161.32 & 557.26 & 623.71 \\
   SDLM ppl & 68.47 & 114.95 & 465.29 & 476.45 \\
   $\Delta$ppl & 3.09 & 16.36 & 91.98 & 147.25 \\
   $\Delta$ppl/Baseline ppl & 4.3\% & 10.1\% & 16.5\% & 23.61\%\\
   \hline
   \hline
\end{tabular}
\caption{Perplexity of words with different mean number of sememes on the test set.}
\label{mean_sememe_tb}
\end{table}

We also test the robustness of our model by randomly removing 10\% sememe-sense connections in HowNet. The test perplexity for Tied LSTM (medium) + SDLM slightly goes up to 97.67, compared to 97.32 with a complete HowNet, which shows that our model is robust to tiny incompleteness of annotations. However, the performance of out model is still largely dependent upon the accuracy of sememe annotations. As HowNet is continuously updated, we expect our model to perform better with sememe knowledge of higher quality. 

\subsection{Headline Generation}

\subsubsection*{Dataset}
We use the LCSTS dataset to evaluate our SDLM structure as the decoder of the sequence-to-sequence model. As its author suggests, we divide the dataset into the training set, the validation set and the test set, whose sizes are 2.4M, 8.7k and 725 respectively. Details can be found in Appendix \ref{suc:preprocess_hg}.

\subsubsection*{Models}

For this task, we consider two models for comparison.

\textbf{RNN-context}
As described in \citep{bahdanau2014neural}, RNN-context is a basic sequence-to-sequence model with a bi-LSTM encoder, an LSTM decoder and attention mechanism adopted. The context vector is concatenated with the word embedding at each timestep when decoding. It's widely used for sequence-to-sequence learning, so we set it as the baseline model.

\textbf{RNN-context-SDLM}
Based on RNN-context, we substitute the decoder with our proposed SDLM and name it RNN-context-SDLM.

\subsubsection*{Experimental Settings}
We implement our models with PyTorch, on top of the OpenNMT libraries\footnote{\url{http://opennmt.net}}. For both models, we set the word embedding size to 250, the hidden unit size to 250, the vocabulary size to 40000, and the beam size of the decoder to 5. For RNN-context-SDLM, we set the number of basis matrices to 3. We conduct a hyper-parameter search for both models (see Appendix \ref{sec:search_hg} for settings and optimal hyper-parameters). 

\subsubsection*{Experimental Results}
Following previous works, we report the F1-score of ROUGE-1, ROUGE-2, and ROUGE-L on the test set. Table \ref{rouge_score} shows that our model outperforms the baseline model on all metrics. We attribute the improvement to the use of SDLM structure. 

Words in headlines do not always appear in the corresponding articles. However, words with the same sememes have a high probability to appear in the articles intuitively. Therefore, a probable reason for the improvement is that our model could predict sememes highly relevant to the article, thus generate more accurate headlines. This could be corroborated by our case study.

\begin{table}[!htb]
\centering \small
\begin{tabular}{lrrr}
  \hline
  \hline
  Model & Rouge-1 & Rouge-2 & Rouge-L\\
  \hline
  RNN-context & 38.2 & 25.7 & 35.4 \\
  RNN-context-SDLM & \textbf{38.8} & \textbf{26.2} & \textbf{36.1} \\
  \hline
  \hline
\end{tabular}
\caption{ROUGE scores of both models on the LCSTS test set. }
\label{rouge_score}
\end{table}

\subsection{Case Study}
\label{sec:case study}
\begin{table}[!b]
\centering \small
\resizebox{\linewidth}{!}{
\begin{tabular}{lll}
  \hline
  \hline
  \multicolumn{3}{c}{Example (1)}\\
  \multicolumn{3}{l}{\begin{CJK}{UTF8}{gbsn}去年 美国 贸易逆差 初步 估计 为 $<$N$>$  \underline {\hbox to 10mm{}} 。\end{CJK}}\\
  \multicolumn{3}{l}{The U.S. trade deficit last year is initially estimated to be $<$N$>$ \underline {\hbox to 10mm{}} .}\\
  \hline
  \multicolumn{3}{c}{Top 5 word prediction} \\
  \begin{CJK}{UTF8}{gbsn}\textbf{美元}\end{CJK}\textbf{ ``dollar''} & \begin{CJK}{UTF8}{gbsn}，\end{CJK} ``,'' & \begin{CJK}{UTF8}{gbsn}。\end{CJK} ``.''\\
  \begin{CJK}{UTF8}{gbsn}日元\end{CJK} ``yen'' & \begin{CJK}{UTF8}{gbsn}和\end{CJK} ``and'' & \\
  \hline
  \multicolumn{3}{c}{Top 5 sememe prediction} \\
  \begin{CJK}{UTF8}{gbsn}\textbf{商业}\end{CJK}\textbf{ ``commerce''} & \begin{CJK}{UTF8}{gbsn}\textbf{金融}\end{CJK}\textbf{ ``finance''} & \begin{CJK}{UTF8}{gbsn}\textbf{单位}\end{CJK}\textbf{ ``unit''} \\
  \begin{CJK}{UTF8}{gbsn}多少\end{CJK} ``amount'' & \begin{CJK}{UTF8}{gbsn}专\end{CJK} ``proper name'' & \\
  \hline
  \hline
  \multicolumn{3}{c}{Example (2)}\\
  \multicolumn{3}{l}{\begin{CJK}{UTF8}{gbsn}阿 总理 \underline {\hbox to 10mm{}} 已 签署 了 一 项 命令 。\end{CJK}}\\
  \multicolumn{3}{l}{Albanian Prime Minister \underline {\hbox to 10mm{}} has signed an order.}\\
  \hline
  \multicolumn{3}{c}{Top 5 word prediction} \\
  \begin{CJK}{UTF8}{gbsn}内\end{CJK} ``inside'' & \textbf{$\bm{<}$unk$\bm{>}$} & \begin{CJK}{UTF8}{gbsn}在\end{CJK} ``at'' \\
  \begin{CJK}{UTF8}{gbsn}塔\end{CJK} ``tower'' & \begin{CJK}{UTF8}{gbsn}和\end{CJK} ``and'' & \\
  \hline
  \multicolumn{3}{c}{Top 5 sememe prediction} \\
  \begin{CJK}{UTF8}{gbsn}\textbf{政}\end{CJK}\textbf{ ``politics''} & \begin{CJK}{UTF8}{gbsn}\textbf{人}\end{CJK}\textbf{ ``person''} & \begin{CJK}{UTF8}{gbsn}花草\end{CJK} ``flowers'' \\
  \begin{CJK}{UTF8}{gbsn}\textbf{担任}\end{CJK}\textbf{ ``undertake''} & \begin{CJK}{UTF8}{gbsn}水域\end{CJK} ``waters'' & \\
  \hline
  \hline
\end{tabular}}
\caption{Some examples of word and sememe predictions on the test set of the People's Daily Corpus. }
\label{lm_case_study}
\end{table}	

The above experiments demonstrate the effectiveness of our SDLM. Here we present some samples from the test set of the People's Daily Corpus in Table \ref{lm_case_study} as well as the LCSTS dataset in Table \ref{hg_case_study} and conduct further analysis. 

For each example of language modeling, given the context of previous words, we list the Top 5 words and Top 5 sememes predicted by our SDLM. The target words and the sememes annotated with them in HowNet are blackened. Note that if the target word is an out-of-vocabulary (OOV) word, helpful sememes that are related to the target meaning are blackened.

Sememes annotated with the corresponding sense of the target word \begin{CJK}{UTF8}{gbsn}美元\end{CJK} ``dollar'' are \begin{CJK}{UTF8}{gbsn}单位\end{CJK} ``unit'', \begin{CJK}{UTF8}{gbsn}商业\end{CJK} ``commerce'', \begin{CJK}{UTF8}{gbsn}金融\end{CJK} ``finance'', \begin{CJK}{UTF8}{gbsn}货币\end{CJK} ``money'' and \begin{CJK}{UTF8}{gbsn}美国\end{CJK} ``US''. In Example (1), the target word  ``dollar'' is predicted correctly and most of its sememes are activated in the predicting process. It indicates that our SDLM has learned the word-sense-sememe hierarchy and used sememe knowledge to improve language modeling. 

Example (2) shows that our SDLM can provide interpretable results on OOV word prediction with sememe information associated with it. The target word here should be the name of the Albanian prime minister, which is out of vocabulary. But with our model, one can still conclude that this word is probably relevant to the sememe ``politics'', ``person'', ``flowers'', ``undertake'' and ``waters'', most of which characterize the meaning of this OOV word -- the name of a politician. This feature can be helpful when the vocabulary size is limited or there are many terminologies and names in the corpus.

For the example of headline generation, given the article and previous words, when generating the word \begin{CJK}{UTF8}{gbsn}生\end{CJK} ``student'', except the sememe \begin{CJK}{UTF8}{gbsn}预料\end{CJK} ``predict'', all other Top 5 predicted sememes have high relevance to either the predicted word or the context. To be specific, the sememe \begin{CJK}{UTF8}{gbsn}学习\end{CJK} ``study'' is annotated with \begin{CJK}{UTF8}{gbsn}生\end{CJK} ``student'' in HowNet. \begin{CJK}{UTF8}{gbsn}考试\end{CJK} ``exam'' indicates ``college entrance exam''. \begin{CJK}{UTF8}{gbsn}特定牌子\end{CJK} ``brand'' indicates ``BMW''. And \begin{CJK}{UTF8}{gbsn}高等\end{CJK} ``higher'' indicates ``higher education'', which is the next step after this exam. We can conclude that with sememe knowledge, our SDLM structure can extract critical information from both the given article and generated words explicitly and produce better summarization based on it.
\begin{table}
\small
\centering
\begin{tabular}{p{7cm}}
  \hline
  \hline
  \multicolumn{1}{c}{Article}\\
\begin{CJK}{UTF8}{gbsn}8 日 ， 阜 新 一 开 宝马 轿车 参加 高考 的 男 考生 考场 作弊 被 抓 ， 因 不满 监考 老师 没收 作弊 手机 ， 从 背后 一 脚 将 女 监考 老师 从 最后 一 排 踹 到 讲台 。 并 口 出 狂 言 ： “ 你 知道 我 爸 是 谁 啊 ， 你 就 查 我 ？ ” 目前 ， 打人 考生 已 被拘留 。\end{CJK}\\
  On the 8th in Fuxin, a male student drove a BMW to take the college entrance exam and was caught cheating. Because the teacher confiscated his mobile phone, he kicked the teacher from the last row to the podium and shouted: "Do you know who my dad is? How dare you catch me!" Currently, this student has been detained.\\
  \hline
  \multicolumn{1}{c}{Gold}\\
  \begin{CJK}{UTF8}{gbsn}男生 高考 作弊 追打 监考 老师 ：你 知道 我 爸 是 谁 ？\end{CJK}\\
  In the college entrance exam, a male student caught cheating hit the teacher: Do you know who my dad is?\\
  \hline
  \multicolumn{1}{c}{RNN-context-SDLM}\\
 \begin{CJK}{UTF8}{gbsn}高考 \underline{生} 作弊 被 抓 ：你 知道 我 爸 是 谁 啊 ？\end{CJK}\\
  In the college entrance exam, a \underline{student} was caught cheating: Do you know who my dad is?\\
  \hline  
  \multicolumn{1}{c}{Top 5 sememe prediction}\\
  \begin{CJK}{UTF8}{gbsn}\textbf{考试 ``exam'' } \end{CJK} \quad \begin{CJK}{UTF8}{gbsn}\textbf{学习 ``study''} \end{CJK} \quad \begin{CJK}{UTF8}{gbsn}\textbf{特定牌子 ``brand''} \end{CJK}\\
  \begin{CJK}{UTF8}{gbsn}预料 ``predict'' \end{CJK}  \quad \begin{CJK}{UTF8}{gbsn}\textbf{高等 ``higher''}\end{CJK}\\
  \hline
  \hline
\end{tabular}
\caption{An example of generated headlines on the LCSTS test set. }
\label{hg_case_study}
\end{table}	

%% file: relatedwork.tex
\section{Related Work}

\paragraph{Neural Language Modeling.} RNNs have achieved state-of-the-art performance in the language modeling task since \citet{mikolov2010recurrent} first apply RNNs for language modeling. Much work has been done to improve RNN-based language modeling. For example, a variety of work \citep{rnn-dropout,variational-lstm,Merity2017Revisiting,awd-lstm-lm} introduces many regularization and optimization methods for RNNs. Based on the observation that the word appearing in the previous context is more likely to appear again, some work \citep{Grave2017Unbounded, Grave2016Improving} proposes to use cache for improvements. In this paper, we mainly focus on the output decoder, the module between the context vector and the predicted probability distribution. Similar to our SDLM, \citet{Yang2017Breaking} propose a high-rank model which adopts a Mixture of Softmaxes structure for the output decoder. However, our model is sememe-driven with each expert corresponding to an interpretable sememe.

\paragraph{Hierarchical Decoder} Since softmax computation on large vocabulary is time-consuming, therefore being a dominant part of the model's complexity, various hierarchical softmax models have been proposed to address this issue. These models can be categorized to class-based models and tree-based models according to their hierarchical structure. \citet{Goodman2001Classes} first proposes the class-based model which divides the whole vocabulary into different classes and uses a hierarchical softmax decoder to model the probability as $\mathbb{P}(\text{word})=\mathbb{P}(\text{word}|\text{class})\mathbb{P}(\text{class})$, which is similar to our model. For the tree-based models, all words are organized in a tree structure and the word probability is calculated as the probability of always choosing the correct child along the path from the root node to the word node. While \citet{Morin2005Hierarchical} utilize knowledge from WordNet to build the tree, \citet{Mnih2008A} build it in a bootstrapping way and \citet{Mikolov2013Distributed} construct a Huffman Tree based on word frequencies. Recently, \citet{jiangexploration} reform the tree-based structure to make it more efficient on GPUs. The major differences between our model and theirs are the purpose and the motivation. Our model targets at improving the performance and interpretability of language modeling using external knowledge in HowNet. Therefore, we take its philosophy of the word-sense-sememe hierarchy to design our hierarchical decoder. Meanwhile, the class-based and tree-based models are mainly designed to speed up the softmax computation in the training process.

\paragraph{Sememe.} Recently, there are a lot of works concentrating on utilizing sememe knowledge in traditional natural language processing tasks. For example, \citet{niu2017improved} use sememe knowledge to improve the quality of word embeddings and cope with the problem of word sense disambiguation. \citet{xie2017lexical} apply matrix factorization to predict sememes for words. \citet{P18-1227} improve their work by incorporating character-level information. Our work extends the previous works and tries to combine word-sense-sememe hierarchy with the sequential model. To be specific, this is the first work to improve the performance and interpretability of Neural Language Modeling with sememe knowledge.

\paragraph{Product of Experts.} As \citet{Hinton1999Products, Hinton2002Training} propose, the final probability can be calculated as the product of probabilities given by experts.\citet{Gales2006Product} apply PoE to the speech recognition where each expert is a Gaussian mixture model. Unlike their work, in our SDLM, each expert is mapped to a sememe with better interpretability. Moreover, as the final distribution is a categorical distribution, each expert is only responsible for making predictions on a subset of the categories (usually less than 10), so we call it Sparse Product of Experts.

\paragraph{Headline Generation.} Headline generation is a kind of text summarization tasks. In recent years, with the advances of RNNs, a lot of works have been done in this domain. The encoder-decoder models \citep{sutskever2014sequence,cho2014learning} have achieved great success in sequence-to-sequence learning. \citet{rush2015neural} propose a local attention-based model for abstractive sentence summarization. \citet{gu2016incorporating} introduce the copying mechanism which is close to the rote memorization of the human being. \citet{ayana2016neural} employ the minimum risk training strategy to optimize model parameters. Different from these works, we focus on the decoder of the sequence-to-sequence model, and adopt SDLM to utilize sememe knowledge for sentence generation.

%% file: conclusion.tex
\section{Conclusion and Further Work}

In this paper, we propose an interpretable Sememe-Driven Language Model with a hierarchical sememe-sense-word decoder. Besides interpretability, our model also achieves state-of-the-art performance in the Chinese Language Modeling task and shows improvement in the Headline Generation task. These results indicate that SDLM can successfully take advantages of sememe knowledge.

As for future work, we plan the following research directions: (1) In language modeling, given a sequence of words, a sequence of corresponding sememes can also be obtained.  We will utilize the context sememe information for better sememe and word prediction. (2) Structural information about sememes in HowNet is ignored in our work. We will extend our model with the hierarchical sememe tree for more accurate relations between words and their sememes. (3) It is imaginable that the performance of SDLM will be significantly influenced by the annotation quality of sememe knowledge. We will also devote to further enrich the sememe knowledge for new words and phrases, and investigate its effect on SDLM.

%% file: acknowledgement.tex
\section*{Acknowledgement}
This work is supported by the 973 Program (No. 2014CB340501), the National Natural Science Foundation of China (NSFC No. 61572273) and the research fund of Tsinghua University-Tencent Joint Laboratory for Internet Innovation Technology. This work is also funded by China Association for Science and Technology (2016QNRC001). Hao Zhu and Jun Yan are supported by Tsinghua University Initiative Scientific Research Program. We thank all members of Tsinghua NLP lab. We also thank anonymous reviewers for their careful reading and their insightful comments.

%% file: appendix.tex
\section{Details about Preprocessing of the People's Daily Dataset}

\label{suc:preprocess_lm}

In this section, we describe the details about preprocessing of the People's Daily dataset.

Firstly, we treat sentence, which is segmented by particular punctuations, as the minimum unit and then shuffle the corpus. We split the corpus into the training set, the validation set and the test set, which contain 734k, 10k, 19k words respectively. Similar to the preprocessing performed in \citep{mikolov2010recurrent}, we replace the number with \textless N\textgreater, the specific date with \textless date\textgreater, the year with \textless year\textgreater, and the time with \textless time\textgreater. Different from the preprocessing of the English language modeling dataset, we keep the punctuations and therefore do not append \textless eos\textgreater $ $ at the end of each sentence. Those words that occur less than 5 times are replaced with \textless unk\textgreater.

Since our model requires that every word should be included in the dictionary of HowNet, we segment each non-annotated word into annotated words with the forward maximum matching algorithm. 

\section{Details about Preprocessing of the LCSTS Dataset}

\label{suc:preprocess_hg}

In this section, we describe the details about preprocessing of the LCSTS dataset.

The dataset consists of over 2 million article-headline pairs collected from Sina Weibo, the most popular social media network in China. It’s composed of three parts. Each pair from PART-II and PART-III is labeled with a score which indicates the relevance between the article and its headline. As its author suggests, we take pairs from a subset of PART-II as the validation set and a subset of PART-III as the test set. Only pairs with score 3, 4 and 5, which means high relevance, are taken into account. We take pairs from PART-I that do not occur in the validation set as the training set.

Similar to what we do for preprocessing the People's Daily dataset, the word segmentation is carried out with jieba\footnote{\url{https://pypi.python.org/pypi/jieba}} based on the dictionary of HowNet to alleviate the OOV problems.

\section{Details about Experiments Setting}

In this section we describe the strategy we adopt to choose hyper-parameters and the optimal hyper-parameters used in the experiment.

\subsection{Language Modeling}

\label{sec:search_lm}
The hyper-parameters are chosen according to the performance on the validation set. For medium (Tied) LSTM and its cHSM, tHSM variants, we search the dropout rate from $\{0.45, 0.5, 0.55, 0.6, 0.65, 0.7\}$. For large (Tied) LSTM and its cHSM, tHSM variants, we search the dropout rate from $\{0.6, 0.65, 0.7, 0.75, 0.8\}$. For AWD-LSTM and its variants, we follow most of the hyper-parameters described in \citep{awd-lstm-lm} and only search the dropout rates (embedding V-dropout from $\{0.35, 0.4, 0.45, 0.5\}$, hidden state V-dropout from $\{0.2, 0.25, 0.3\}$, word level V-dropout from $\{0.05, 0.1, 0.15\}$ and context vector dropout from $\{0.4, 0.5\}$). For our SDLM and MoS, we fix all other hyper-parameters and only search the dropout rates of the last two layers respectively from \{0.35, 0.4, 0.45\} and \{0.25, 0.3\}. The initial learning rate for MoS on the top of AWD-LSTM is set to 20 to avoid diverging.

For (Tied) LSTM, we set the hidden unit and word embedding size to 650 (medium) / 1500 (large), batch size to 20, bptt to 35, dropout rate to 0.6 (medium) / 0.7 (large) and initial learning rate to 20. The optimal dropout rates for cHSM and tHSM are 0.55 (cHSM, medium), 0.5 (cHSM, medium, tied), 0.7 (cHSM, large), 0.65 (cHSM, large, tied), 0.55 (tHSM, medium) and 0.7 (tHSM, large). For AWD-LSTM and its variants, the hyper-parameters for the baseline are summarized in Table \ref{hp-awd-lstm}.

\begin{table}
\centering \small
\begin{tabular}{lll}
  \hline
  Hyper-parameter & Baseline \\
  \hline
  Learning rate & 30 \\
  Batch size & 15 \\
  Embedding size & 400 \\
  RNN hidden size & [1150, 1150, 400] \\
  Word-level V-dropout & 0.1 \\
  Embedding V-dropout & 0.5 \\
  Hidden state V-dropout & 0.2 \\
  Recurrent weight dropout & 0.5 \\
  Context vector dropout & 0.4 \\
  \hline
\end{tabular}
\caption{Hyper-parameters used for AWD-LSTM and its variants}
\label{hp-awd-lstm}
\end{table}

\subsection{Headline Generation}

\label{sec:search_hg}
The hyper-parameters are chosen according to the performance on the validation set. For RNN-context, we search the dropout rate from $\{0.1, 0.15, 0.2, 0.25, 0.3, 0.35\}$, the batch size from $\{32, 64\}$ and try SGD and Adam optimizers. For RNN-context-SDLM, we search the dropout rate from $\{0.15, 0.2, 0.25\}$, the batch size from $\{32, 64\}$ and try SGD and Adam optimizers.

For RNN-context, we use Adam optimizer with starting learning rate 0.001. The batch size is 32 and the dropout rate is 0.15. For RNN-context-SDLM, we use Adam optimizer with starting learning rate 0.001. The batch size is 64 and the dropout rate is 0.2. 

%% file: emnlp2018.bbl
\begin{thebibliography}{43}
\expandafter\ifx\csname natexlab\endcsname\relax\def\natexlab#1{#1}\fi

\bibitem[{Ayana et~al.(2016)Ayana, Liu, and Sun}]{ayana2016neural}
Shiqi~Shen Ayana, Zhiyuan Liu, and Maosong Sun. 2016.
\newblock Neural headline generation with minimum risk training.
\newblock \emph{arXiv preprint arXiv:1604.01904}.

\bibitem[{Bahdanau et~al.(2015)Bahdanau, Cho, and Bengio}]{bahdanau2014neural}
Dzmitry Bahdanau, Kyunghyun Cho, and Yoshua Bengio. 2015.
\newblock Neural machine translation by jointly learning to align and
  translate.
\newblock In \emph{Proceedings of ICLR}.

\bibitem[{Banko et~al.(2000)Banko, Mittal, and Witbrock}]{banko2000headline}
Michele Banko, Vibhu~O Mittal, and Michael~J Witbrock. 2000.
\newblock Headline generation based on statistical translation.
\newblock In \emph{Proceedings of ACL}, pages 318--325. Association for
  Computational Linguistics.

\bibitem[{Bengio et~al.(2003)Bengio, Ducharme, Vincent, and Jauvin}]{neural-lm}
Yoshua Bengio, Rejean Ducharme, Pascal Vincent, and Christian Jauvin. 2003.
\newblock {A neural probabilistic language model}.
\newblock \emph{Journal of Machine Learning Research}, 3:1137--1155.

\bibitem[{Berger and Lafferty(1999)}]{berger1999information}
Adam Berger and John Lafferty. 1999.
\newblock Information retrieval as statistical translation.
\newblock In \emph{Proceedings of SIGIR}, pages 222--229. ACM.

\bibitem[{Brants et~al.(2007)Brants, Popat, Xu, Och, and
  Dean}]{brants2007large}
Thorsten Brants, Ashok~C Popat, Peng Xu, Franz~J Och, and Jeffrey Dean. 2007.
\newblock Large language models in machine translation.
\newblock In \emph{Proceedings of EMNLP}.

\bibitem[{Brown et~al.(1990)Brown, Cocke, Pietra, Pietra, Jelinek, Lafferty,
  Mercer, and Roossin}]{brown1990statistical}
Peter~F Brown, John Cocke, Stephen A~Della Pietra, Vincent J~Della Pietra,
  Fredrick Jelinek, John~D Lafferty, Robert~L Mercer, and Paul~S Roossin. 1990.
\newblock A statistical approach to machine translation.
\newblock \emph{Computational linguistics}, 16(2):79--85.

\bibitem[{Cho et~al.(2014)Cho, Van~Merri{\"e}nboer, Gulcehre, Bahdanau,
  Bougares, Schwenk, and Bengio}]{cho2014learning}
Kyunghyun Cho, Bart Van~Merri{\"e}nboer, Caglar Gulcehre, Dzmitry Bahdanau,
  Fethi Bougares, Holger Schwenk, and Yoshua Bengio. 2014.
\newblock Learning phrase representations using rnn encoder-decoder for
  statistical machine translation.
\newblock In \emph{Proceedings of EMNLP}.

\bibitem[{Dong and Dong(2006)}]{zhendong2006hownet}
Zhendong Dong and Qiang Dong. 2006.
\newblock \emph{Hownet and the computation of meaning (with Cd-rom)}.
\newblock World Scientific.

\bibitem[{Fu et~al.(2013)Fu, Liu, Guo, and Wang}]{xianghua2013multi}
Xianghua Fu, Guo Liu, Yanyan Guo, and Zhiqiang Wang. 2013.
\newblock Multi-aspect sentiment analysis for chinese online social reviews
  based on topic modeling and hownet lexicon.
\newblock \emph{Knowledge-Based Systems}, 37:186--195.

\bibitem[{Gal and Ghahramani(2016)}]{variational-lstm}
Yarin Gal and Zoubin Ghahramani. 2016.
\newblock {A theoretically grounded application of dropout in recurrent neural
  networks}.
\newblock In \emph{Proceedings of NIPS}.

\bibitem[{Gales and Airey(2006)}]{Gales2006Product}
M.~J.~F. Gales and S.~S. Airey. 2006.
\newblock Product of gaussians for speech recognition.
\newblock \emph{Computer Speech and Language}, 20(1):22--40.

\bibitem[{Goodman(2001)}]{Goodman2001Classes}
J~Goodman. 2001.
\newblock Classes for fast maximum entropy training.
\newblock In \emph{Proceedings of ICASSP}, pages 561--564 vol.1.

\bibitem[{Grave et~al.(2017{\natexlab{a}})Grave, Cisse, and
  Joulin}]{Grave2017Unbounded}
Edouard Grave, Moustapha Cisse, and Armand Joulin. 2017{\natexlab{a}}.
\newblock Unbounded cache model for online language modeling with open
  vocabulary.
\newblock In \emph{Proceedings of NIPS}.

\bibitem[{Grave et~al.(2017{\natexlab{b}})Grave, Joulin, and
  Usunier}]{Grave2016Improving}
Edouard Grave, Armand Joulin, and Nicolas Usunier. 2017{\natexlab{b}}.
\newblock Improving neural language models with a continuous cache.
\newblock In \emph{Proceedings of ICLR}.

\bibitem[{Gu et~al.(2016)Gu, Lu, Li, and Li}]{gu2016incorporating}
Jiatao Gu, Zhengdong Lu, Hang Li, and Victor~OK Li. 2016.
\newblock Incorporating copying mechanism in sequence-to-sequence learning.
\newblock In \emph{Proceedings of ACL}, pages 1631--1640.

\bibitem[{Hiemstra(1998)}]{hiemstra1998linguistically}
Djoerd Hiemstra. 1998.
\newblock A linguistically motivated probabilistic model of information
  retrieval.
\newblock In \emph{Proceedings of TPDL}, pages 569--584. Springer.

\bibitem[{Hinton(1999)}]{Hinton1999Products}
G.~E Hinton. 1999.
\newblock Products of experts.
\newblock In \emph{Artificial Neural Networks, 1999. ICANN 99. Ninth
  International Conference on}, pages 1--6 vol.1.

\bibitem[{Hinton(2002)}]{Hinton2002Training}
G.~E. Hinton. 2002.
\newblock \emph{Training products of experts by minimizing contrastive
  divergence.}
\newblock MIT Press.

\bibitem[{Hu et~al.(2015)Hu, Chen, and Zhu}]{hu2015lcsts}
Baotian Hu, Qingcai Chen, and Fangze Zhu. 2015.
\newblock Lcsts: A large scale chinese short text summarization dataset.
\newblock In \emph{Proceedings of EMNLP}.

\bibitem[{Inan et~al.(2017)Inan, Khosravi, and Socher}]{weight-tying}
Hakan Inan, Khashayar Khosravi, and Richard Socher. 2017.
\newblock {Tying word vectors and word classifiers: A loss framework for
  language modeling}.
\newblock In \emph{Proceedings of ICLR}.

\bibitem[{Jiang et~al.(2017)Jiang, Rong, Gao, Shen, Xiong, Jiang, Rong, Gao,
  Shen, and Xiong}]{jiangexploration}
Nan Jiang, Wenge Rong, Min Gao, Yikang Shen, Zhang Xiong, Nan Jiang, Wenge
  Rong, Min Gao, Yikang Shen, and Zhang Xiong. 2017.
\newblock Exploration of tree-based hierarchical softmax for recurrent language
  models.
\newblock In \emph{Proceedings of IJCAI}.

\bibitem[{Jin et~al.(2018)Jin, Zhu, Liu, Xie, Sun, Lin, and Lin}]{P18-1227}
Huiming Jin, Hao Zhu, Zhiyuan Liu, Ruobing Xie, Maosong Sun, Fen Lin, and Leyu
  Lin. 2018.
\newblock Incorporating chinese characters of words for lexical sememe
  prediction.
\newblock In \emph{Proceedings of ACL}, pages 2439--2449. Association for
  Computational Linguistics.

\bibitem[{Jurafsky(2000)}]{jurafsky2000speech}
Dan Jurafsky. 2000.
\newblock Speech \& language processing.
\newblock chapter~4. Pearson Education India.

\bibitem[{Katz(1987)}]{katz1987estimation}
Slava Katz. 1987.
\newblock Estimation of probabilities from sparse data for the language model
  component of a speech recognizer.
\newblock \emph{IEEE transactions on acoustics, speech, and signal processing},
  35(3):400--401.

\bibitem[{Liu(2002)}]{liu2002word}
Qun Liu. 2002.
\newblock Word similarity computing based on hownet.
\newblock \emph{Computational linguistics and Chinese language processing},
  7(2):59--76.

\bibitem[{Marcus et~al.(1993)Marcus, Santorini, and Marcinkiewicz}]{penn}
Mitchell~P. Marcus, Beatrice Santorini, and Ann Marcinkiewicz, Mary. 1993.
\newblock {Building a large annotated corpus of English: The Penn Treebank}.
\newblock \emph{Computational Linguistics}, 19:313--330.

\bibitem[{Merity et~al.(2018)Merity, Keskar, and Socher}]{awd-lstm-lm}
Stephen Merity, Nitish~Shirish Keskar, and Richard Socher. 2018.
\newblock {Regularizing and optimizing LSTM language models}.
\newblock In \emph{Proceedings of ICLR}.

\bibitem[{Merity et~al.(2017)Merity, Mccann, and Socher}]{Merity2017Revisiting}
Stephen Merity, Bryan Mccann, and Richard Socher. 2017.
\newblock Revisiting activation regularization for language rnns.
\newblock \emph{arXiv preprint arXiv:1708.01009}.

\bibitem[{Mikolov et~al.(2010)Mikolov, Karafi{\'a}t, Burget,
  {\v{C}}ernock{\`y}, and Khudanpur}]{mikolov2010recurrent}
Tom{\'a}{\v{s}} Mikolov, Martin Karafi{\'a}t, Luk{\'a}{\v{s}} Burget, Jan
  {\v{C}}ernock{\`y}, and Sanjeev Khudanpur. 2010.
\newblock Recurrent neural network based language model.
\newblock In \emph{Proceedings of INTERSPEECH}.

\bibitem[{Mikolov et~al.(2013)Mikolov, Sutskever, Chen, Corrado, and
  Dean}]{Mikolov2013Distributed}
Tomas Mikolov, Ilya Sutskever, Kai Chen, Greg Corrado, and Jeffrey Dean. 2013.
\newblock Distributed representations of words and phrases and their
  compositionality.
\newblock In \emph{Proceedings of NIPS}, pages 3111--3119.

\bibitem[{Miller et~al.(1999)Miller, Leek, and Schwartz}]{miller1999hidden}
David~RH Miller, Tim Leek, and Richard~M Schwartz. 1999.
\newblock A hidden markov model information retrieval system.
\newblock In \emph{Proceedings of SIGIR}, pages 214--221. ACM.

\bibitem[{Mnih and Hinton(2008)}]{Mnih2008A}
Andriy Mnih and Geoffrey Hinton. 2008.
\newblock A scalable hierarchical distributed language model.
\newblock In \emph{Proceedings of NIPS}, pages 1081--1088.

\bibitem[{Morin and Bengio(2005)}]{Morin2005Hierarchical}
Frederic Morin and Yoshua Bengio. 2005.
\newblock Hierarchical probabilistic neural network language model.
\newblock In \emph{Proceedings of AISTATS}.

\bibitem[{Niu et~al.(2017)Niu, Xie, Liu, and Sun}]{niu2017improved}
Yilin Niu, Ruobing Xie, Zhiyuan Liu, and Maosong Sun. 2017.
\newblock Improved word representation learning with sememes.
\newblock In \emph{Proceedings of ACL}, volume~1, pages 2049--2058.

\bibitem[{Ponte and Croft(1998)}]{ponte1998language}
Jay~M Ponte and W~Bruce Croft. 1998.
\newblock A language modeling approach to information retrieval.
\newblock In \emph{Proceedings of SIGIR}, pages 275--281. ACM.

\bibitem[{Press and Wolf(2017)}]{weight-tying-2}
Ofir Press and Lior Wolf. 2017.
\newblock {Using the output embedding to improve language models}.
\newblock In \emph{Proceedings of EACL}.

\bibitem[{Qi et~al.(2018)Qi, Lin, Sun, Zhu, Xie, and Liu}]{qi2018cross}
Fanchao Qi, Yankai Lin, Maosong Sun, Hao Zhu, Ruobing Xie, and Zhiyuan Liu.
  2018.
\newblock Cross-lingual lexical sememe prediction.
\newblock In \emph{Proceedings of EMNLP}.

\bibitem[{Rush et~al.(2015)Rush, Chopra, and Weston}]{rush2015neural}
Alexander~M Rush, Sumit Chopra, and Jason Weston. 2015.
\newblock A neural attention model for abstractive sentence summarization.
\newblock In \emph{Proceedings of EMNLP}.

\bibitem[{Sutskever et~al.(2014)Sutskever, Vinyals, and
  Le}]{sutskever2014sequence}
Ilya Sutskever, Oriol Vinyals, and Quoc~V Le. 2014.
\newblock Sequence to sequence learning with neural networks.
\newblock In \emph{Proceedings of NIPS}, pages 3104--3112.

\bibitem[{Xie et~al.(2017)Xie, Yuan, Liu, and Sun}]{xie2017lexical}
Ruobing Xie, Xingchi Yuan, Zhiyuan Liu, and Maosong Sun. 2017.
\newblock Lexical sememe prediction via word embeddings and matrix
  factorization.
\newblock In \emph{Proceedings of IJCAI}, pages 4200--4206. AAAI Press.

\bibitem[{Yang et~al.(2018)Yang, Dai, Salakhutdinov, and
  Cohen}]{Yang2017Breaking}
Zhilin Yang, Zihang Dai, Ruslan Salakhutdinov, and William~W. Cohen. 2018.
\newblock Breaking the softmax bottleneck: A high-rank rnn language model.
\newblock In \emph{Proceedings of ICLR}.

\bibitem[{Zaremba et~al.(2014)Zaremba, Sutskever, and Vinyals}]{rnn-dropout}
Wojciech Zaremba, Ilya. Sutskever, and Oriol. Vinyals. 2014.
\newblock {Recurrent neural network regularization}.
\newblock \emph{arXiv preprint arXiv:1409.2329}.

\end{thebibliography}
